%% file: A_dynamical_theory_of_sequential_retrieval_in_input-driven_Hopfield_networks.tex
\documentclass{article}

\usepackage{arxiv}

\usepackage[utf8]{inputenc} 
\usepackage[T1]{fontenc}    
\usepackage{hyperref}       
\usepackage{url}            
\usepackage{booktabs}       
\usepackage{amsfonts}       
\usepackage{nicefrac}       
\usepackage{microtype}      
\usepackage{lipsum}		
\usepackage{graphicx}
\usepackage{natbib}
\usepackage{doi}

\usepackage{times}

\usepackage{bbm}
\usepackage{mathbbol}
\usepackage{imakeidx}

\usepackage{amsmath, amssymb, amsthm}
\usepackage{accents}
\usepackage{upgreek}
\newlength{\dhatheight}

\usepackage{mathtools}

\DeclareMathAlphabet{\mymathbb}{U}{BOONDOX-ds}{m}{n}
\usepackage{yhmath}
\usepackage{mathrsfs}
\usepackage{mathdots}
\usepackage{subfig}
\usepackage[makeroom]{cancel}

\usepackage{tikz}
\usetikzlibrary{matrix, backgrounds, calc}
\usetikzlibrary{graphs, quotes}
\usepackage{geometry}
\usepackage{titlesec}

\newtheorem{remark}{Remark}
\newtheorem{assumption}{Assumption}

\MakeRobust{\eqref}

\definecolor{gnred1}{RGB}{71,0,0} 
\definecolor{gnred2}{RGB}{117,0,0} 
\definecolor{gnred3}{RGB}{164,0,0} 
\definecolor{gnred4}{RGB}{211,0,0} 
\definecolor{gnred5}{RGB}{255,0,0} 
\definecolor{gnred6}{RGB}{255,42,34} 
\definecolor{gnred7}{RGB}{255,91,89} 

\definecolor{gnblue1}{RGB}{0,36,71}   
\definecolor{gnblue2}{RGB}{0,60,118}  
\definecolor{gnblue3}{RGB}{0,85,164}  
\definecolor{gnblue4}{RGB}{0,108,212} 
\definecolor{gnblue5}{RGB}{0,133,255}  
\definecolor{gnblue6}{RGB}{35,156,255} 
\definecolor{gnblue7}{RGB}{88,177,255} 

\definecolor{gnbrown1}{RGB}{71,27,0}  
\definecolor{gnbrown2}{RGB}{117,45,0} 
\definecolor{gnbrown3}{RGB}{164,62,0} 
\definecolor{gnbrown4}{RGB}{211,80,0} 
\definecolor{gnbrown5}{RGB}{255,97,0} 
\definecolor{gnbrown6}{RGB}{255,127,26} 
\definecolor{gnbrown7}{RGB}{255,155,86} 

\hypersetup{colorlinks=true, linkcolor=gnblue4, breaklinks=true, urlcolor=gnblue4, citecolor=gnblue4}

\newcommand{\En}{\mathrm{E}}

\newcommand{\W}{\operatorname{W}}

\newcommand{\1}{\mathbf{1}}

\newcommand{\real}{\mathbb{R}}

\renewcommand{\real}{\mathbb{R}}

\newcommand{\diag}{\mathrm{diag}}

\DeclareSymbolFont{bbold}{U}{bbold}{m}{n}
\DeclareSymbolFontAlphabet{\mathbbold}{bbold}

\newcommand{\vect}[1]{\mathbbold{#1}}
\newcommand{\vectorones}[1][]{\vect{1}_{#1}}

\input{math_commands.tex}

\usepackage{hyperref}
\usepackage{url}

\title{A Dynamical Theory of Sequential Retrieval\\ in Input-Driven Hopfield Networks}

\author{ \href{https://orcid.org/0009-0000-3444-0838}{\includegraphics[scale=0.06]{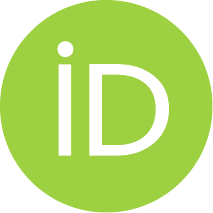}\hspace{1mm}Simone Betteti} \\
	The Italian Institute of\\
	Artificial Intelligence for Industry\\
	Turin, 10129, IT \\
	\texttt{simone.betteti[at]ai4i.it} \\
	\And
	\href{https://orcid.org/0000-0002-9439-296X}{\includegraphics[scale=0.06]{orcid.pdf}\hspace{1mm}Giacomo Baggio} \\
	Department of Information Engineering\\
	Università degli Studi di Padova\\
	Padova, 35131, IT \\
	\texttt{baggio[at]dei.unipd.it} \\
	\AND
    \href{https://orcid.org/0000-0001-8926-1888}{\includegraphics[scale=0.06]{orcid.pdf}\hspace{1mm}Sandro Zampieri} \\
	Department of Information Engineering\\
	Università degli Studi di Padova\\
	Padova, 35131, IT \\
	\texttt{zampi[at]dei.unipd.it}\\ 
}

\renewcommand{\headeright}{}
\renewcommand{\undertitle}{}
\renewcommand{\shorttitle}{Dynamical Theory of Sequential Retrieval in IDP Hopfield Networks}

\hypersetup{
pdftitle={A Dynamical Theory of Sequential Retrieval in Input-Driven Hopfield Networks},
pdfsubject={q-bio.NC, q-bio.QM},
pdfauthor={Simone Betteti, Giacomo Baggio, Sandro Zampieri},
pdfkeywords={Sequential Transitions, Associative memory, Timescale Separation, Dynamical Systems},
}

\begin{document}

\maketitle

\begin{abstract}
Reasoning is the ability to integrate internal states and external inputs in a meaningful and semantically consistent flow. Contemporary machine learning (ML) systems increasingly rely on such sequential reasoning, from language understanding to multi-modal generation, often operating over dictionaries of prototypical patterns reminiscent of associative memory models. Understanding retrieval and sequentiality in associative memory models provides a powerful bridge to gain insight into ML reasoning. While the static retrieval properties of associative memory models are well understood, the theoretical foundations of sequential retrieval and multi-memory integration remain limited, with existing studies largely relying on numerical evidence. This work develops a dynamical theory of sequential reasoning in Hopfield networks. We consider the recently proposed input-driven plasticity (IDP) Hopfield network and analyze a two-timescale architecture coupling fast associative retrieval with slow reasoning dynamics. We derive explicit conditions for self-sustained memory transitions, including gain thresholds, escape times, and collapse regimes. Together, these results provide a principled mathematical account of sequentiality in associative memory models, bridging classical Hopfield dynamics and modern reasoning architectures.
\end{abstract}

\section{Introduction}
Associative memory has been a foundational concept in theoretical neuroscience and machine learning for over four decades. In its classical formulation, memory retrieval is modeled as convergence toward stable equilibria of an energy landscape~\citep{HJJ:82, HJJ:84}, whereby a corrupted cue relaxes to a stored pattern through recurrent dynamics, providing a canonical example of distributed computation in neural networks. A large body of work has focused on characterizing the memory capacity of such models~\citep{ADJ-GH-SH:87b, MR-PE-RE:87, TMV-FMV:88, BS-BG-ZS:24}, revealing a fundamental limitation of the original Hopfield network, whose capacity scales sub-linearly with the number of neurons. Modern Hopfield models~\citep{KD-HJJ:16, KD-HJJ:20} substantially extend this paradigm, achieving exponential memory capacity while retaining an energy-based formulation~\citep{DM-HJ-LM:17}. Beyond capacity, these models admit strong interpretative links to contemporary machine learning architectures, including Transformers~\citep{RH-SB-LJ:21, HB-LY-PB:23} and generative diffusion models~\citep{AL:24}. Despite these advances, modern Hopfield networks largely remain static retrieval systems: dynamics halt upon convergence to a memory, and sequential retrieval requires repeated reinitialization. This limitation contrasts with many cognitive and algorithmic tasks, including those addressed by Transformers, which rely on structured transitions across sequences of memories, and motivates the need for dynamical mechanisms that induce controlled transitions between energy minima to support reasoning over time.

Early work by \citet{KD:86} already identified this limitation and proposed a mechanism for sequential memory retrieval via delayed asymmetric interactions that drive transitions between attractors. This idea has since motivated extensive efforts to model memory transitions, with contemporary work~\citep{SH-KI:86} establishing mean-field results on the gain sustaining sequentiality. Recently, several works have revisited sequential retrieval within the modern Hopfield formalism~\citep{KA-ST-SHT:23, TC-HZV-JA:24, KA-LP-ST-SHT:25}, typically by introducing slow variables or delayed neuronal layers that modulate synaptic couplings over time. Related approaches employ slowly evolving parameters to control attractor switching~\citep{HL-SP-XB:23}. While effective in simulations, these augmented dynamics significantly reduce analytical tractability, obscuring the mechanistic understanding that made the original Hopfield model appealing.

\begin{figure}[h]
\begin{center}
\includegraphics[scale=0.8]{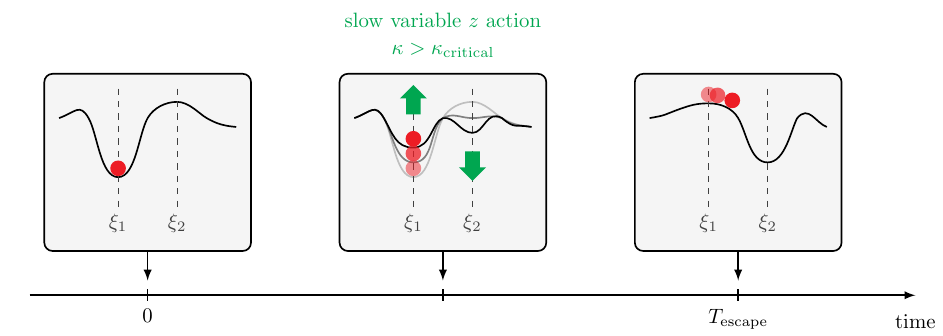}
\end{center}
\caption{\textbf{Schematic illustration of the proposed two-timescale mechanism for sequential memory transitions.} 
At the initial time, the fast state variable is at a stable equilibrium corresponding to memory $\upxi_1$. 
The slow reasoning variable $z$ progressively destabilizes this equilibrium while simultaneously stabilizing the equilibrium associated with memory $\upxi_2$. At the escape time $T_{\mathrm{escape}}$, the equilibrium corresponding to $\upxi_1$ loses stability and the system transitions toward $\upxi_2$. 
This transition occurs only when the gain parameter $\kappa$ of the slow dynamics exceeds a critical threshold $\kappa_{\mathrm{critical}}$. 
We analytically characterize both $k_{\mathrm{critical}}$ and the resulting escape time $T_{\mathrm{escape}}$.}
\label{fig:intro}
\end{figure}

In this paper, we build on recent results on input-controlled transitions in Hopfield networks~\citep{BS-BG-FB-SZ:25idp} to provide a complete dynamical characterization of sequential reasoning in modern Hopfield models. We introduce an input-driven plasticity (IDP) Hopfield framework that exploits a separation of timescales across architectural layers to induce predictable and analytically tractable transitions between memories. This structure enables derivation of exact conditions for memory escape times and gain parameters required to sustain sequential flows. By unifying dynamics and geometry, our framework provides a mechanistic account of how modern Hopfield architectures can implement sequential memory retrieval and support structured reasoning. Figure~\ref{fig:intro} provides a schematic overview of the sequential memory transition mechanism introduced in this work.

\section{Retrieval: the IDP Hopfield model and its modern Hopfield formulation}
Classical associative memory models describe retrieval as a fast relaxation process, whereby a slowly varying external or cortical input cues reconstruction of a stored pattern. In most formulations, however, the explicit dependence of retrieval dynamics on the input is neglected, leading either to sensitivity to initial conditions or to distortions of the stored memories, and limiting algorithmic interpretability. To address this limitation, recent work introduced the input-driven plasticity (IDP) Hopfield model~\citep{BS-BG-FB-SZ:25idp}, an extension of the classical Hopfield network in which synaptic interactions depend multiplicatively on a filtered input. This formulation enables a precise characterization of the input–memory relationship, including equilibrium existence and input-dependent stability.

The IDP Hopfield dynamics are given by
\begin{equation}\label{eq: hopf}
\dot x = -x + W(\upalpha)\Psi(x), \qquad x(0)\in\real^{N},
\end{equation}
where $\upalpha=\upalpha(u)\in\real^{P}$ encodes a filtering of a slow-varying input $u$ and $W(\upalpha)\in\real^{N\times N}$ is a synaptic matrix constructed from stored memories ${\upxi^{1},\dots,\upxi^{P}}\subseteq\real^{N}$. The classical Hopfield model is recovered when all $\{\upalpha_\upmu\}_{\upmu=1}^{P}$ are equal and positive. Details of the construction are deferred to the Appendix.

When the activation $\Psi$ is the gradient of a convex function~\citep{KD-HJJ:20}, the dynamics admit a quasi-gradient representation
\begin{equation}
\dot x = -\frac{\partial\Psi}{\partial x}(x)^{-1}\nabla \En(x,\upalpha),
\end{equation}
and in particular, when $\Psi$ is diagonal, the associated energy becomes 
\begin{equation}\label{eq: ener}
\En(x,\upalpha)
=-\tfrac{1}{2}\Psi(x)^{\top}W(\upalpha)\Psi(x)
+x^{\top}\Psi(x)
-\sum_{i=1}^{N}\int_{0}^{x_{i}}\uppsi(s),ds.
\end{equation}
The energy guarantees convergence to stable equilibria and provides a geometric interpretation of memories and their basins of attraction. Each coefficient $\upalpha_\upnu$ controls the existence and stability of the corresponding memory: the $\upnu$-th memory exists as an equilibrium if and only if $\upalpha_\upnu>\upalpha_{\text{existence}}=1$, and is stable when $\upalpha_\upnu>\upalpha_{\text{stability}}$ exceeds a stability threshold determined by the largest entry of $\upalpha$.

While the IDP Hopfield model captures input-controlled transitions between memories, it does not address the structure of transitions or semantic relations among memories. To this end, we consider a modern Hopfield generalization that introduces a separation of timescales across three interacting layers:
\begin{align}
\uptau_{y}\dot y &= -y + M_{y}\Psi_{y}(x)
&& \boxed{\text{memory layer}},\label{eq: supfast}\\
\uptau_{x}\dot x &= -x + M_{x}[z\odot \Psi_{x}(z\odot y)]
&& \boxed{\text{feature layer}},\label{eq: fast}\\
\uptau_{z}\dot z &= -z + M_{z}\Psi_{z}(u)
&& \boxed{\text{saliency layer}},\label{eq: slow}
\end{align}

Under the timescale separation $\uptau_{z}, \uptau_{x}\gg\uptau_{y}$, the memory layer rapidly indexes stored patterns, the feature layer integrates internal memory structure with reasoning signals, and the slow reasoning layer accumulates evidence from external inputs. This architecture supports continuous, input-modulated memory retrieval and provides the basis for the sequential reasoning dynamics analyzed in the remainder of the paper. Specifically, in the remainder of the paper we will investigate the joint role of feedback modulation $u=x$ and of a sequential transition matrix as first presented in~\citet{KD:86}, and the saliency layer will henceforth be referred to as \emph{reasoning layer}.
    
\section{Reasoning: the two-timescale IDP Hopfield model}
\begin{figure}[!t]
    \centering
    \includegraphics[width=.9\linewidth]{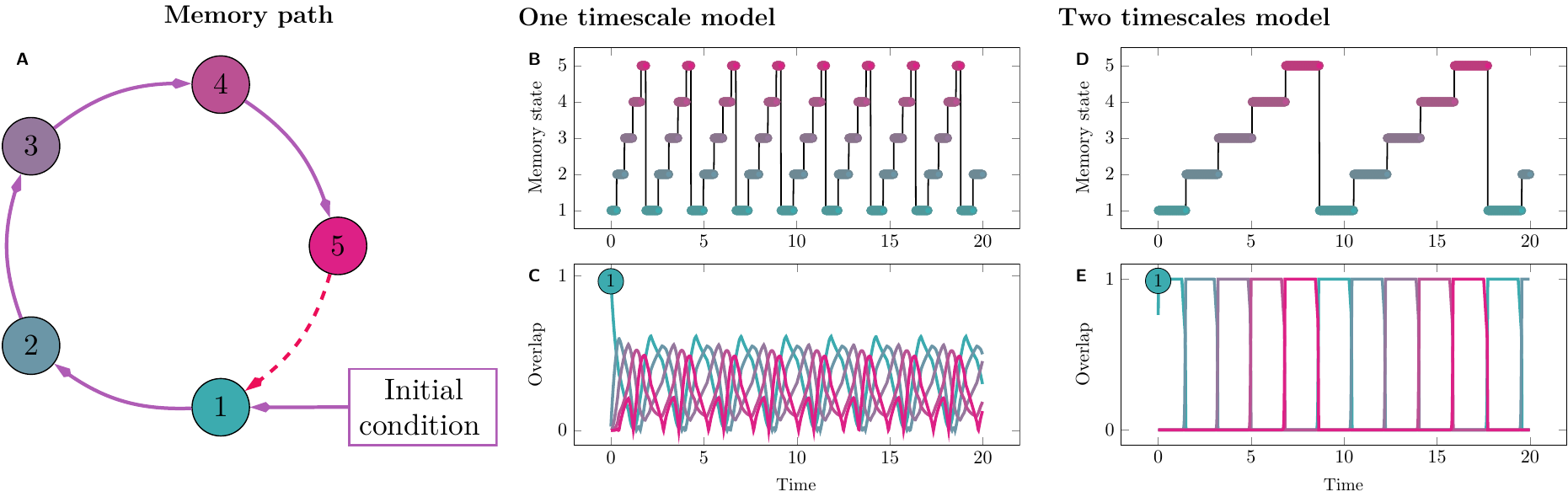}
    \caption{\textbf{Cyclic memory transitions in one- and two-timescale Hopfield models.}
(A) Limit-cycle structure induced by the transition matrices $Q$ in the one-timescale dynamics and $A$ in the two-timescale reasoning dynamics~(\ref{eq: ttreas}), with both systems initialized near memory $\upxi^{1}$. (B-C) One-timescale model: sequential retrieval proceeds through mixed memory states, with partial overlaps and unpredictable escape times across memories. (D-E) Two-timescale reasoning model: transitions occur between individual memories without mixing, with uniform, quantifiable escape times and overlaps reaching their maximal value, indicating exact memory alignment.}
    \label{fig: limitc-noinp}
\end{figure}

Most implementations of modern Hopfield models exploit a timescale separation between memory and feature dynamics to recover a single-timescale system that instantiates different retrieval mechanisms depending on the choice of activation function, e.g. Transformer-like retrieval for $\Psi_{x}(x)=\mathrm{softmax}(x)$~\citep{RH-SB-LJ:21}. Under this limit, the model reduces to a standard associative memory system. In our tripartite architecture, we retain in addition a slow reasoning variable to construct a minimal two-timescale Hopfield model capable of autonomous sequential retrieval. We specialize our model~(\ref{eq: supfast}), (\ref{eq: fast}), (\ref{eq: slow}) to the choices $\uptau_{y}\to 0$, $M_{y}^{\top}=M_{z}^{\top}=M_{x}=M=N^{-1/2}[\upxi^{1},\dots,\upxi^{P}]$, $\Psi_{x}=\mathrm{Id}$ and $\Psi_{y}=\Psi_{z}=\Psi$. Then the reasoning Hopfield model is defined by the coupled dynamics
\begin{equation}\label{eq: ttreas}
\begin{cases}
\uptau_x \dot x = -x + M \diag(\upalpha) M^{\top}\Psi(x), \qquad x(0)\in\real^N,\\
\uptau_z \dot z = -z + \frac{\kappa}{\sqrt{N}} A M^{\top}\Psi(x), \qquad z(0)\in\real^P,\
\upalpha = z \odot z,
\end{cases}
\end{equation}
where $\kappa>0$ is a gain parameter and $\Psi(x)=(\uppsi(x_{1}),\cdots,\uppsi(x_{N}))$ is a diagonal, non-decreasing and saturating activation function. In the following, we consider the HardTanh activation function $\uppsi(z)=\max\{-1,\ \min\{x,1\}\}$. The sequential associations between memories are instead coded by $A\in\real^{P\times P}$, which we name \emph{reasoning matrix}. In our case $A$ is a circulant matrix, i.e., $A_{\upnu\upmu}=\updelta_{\upnu+1,\upnu}$, where $\updelta_{i,j}$ denotes the Kronecker delta, and induces a limit cycle over the memories (see Fig.~\ref{fig: limitc-noinp}A). The circulant matrix $A$ plays a role analogous to the transition matrix
\begin{align}
    Q&=\frac{1}{N}\sum_{\upmu=1}^{P-1}\upxi^{\upmu+1}{\upxi^{\upmu}}^{\top}+\frac{1}{N}\upxi^{1}{\upxi^{P}}^{\top} \in\real^{N\times N}\\
    &=MAM^{\top}
\end{align}
introduced in the seminal work of~\citet{KD:86}, while operating at the level of memory indices rather than feature space. Comparing the dynamics in~(\ref{eq: ttreas}) with the Kleinfeld-type, one timescale Hopfield model $\dot x = -x +\kappa Q\Psi(x)$, we observe important differences in the sequential retrieval. 
As shown in Fig.~\ref{fig: limitc-noinp}B-C, the one-timescale model induces oscillatory behavior across memory components, but exhibits mixed states with low overlap, irregular escape times, and strong sensitivity to the gain $\kappa$. Depending on $\kappa$, the dynamics may collapse to the origin, converge to a fixed point, or sustain a limit cycle, consistent with previous observations~\citep{GM-BN:24}. In contrast, in Fig.~\ref{fig: limitc-noinp}D-E the two-timescale reasoning model displays robust and predictable sequential retrieval for all $\kappa>4$. Transitions are sharp, overlaps reach their maximal value, and escape times are uniform across memories, yielding clean transitions with exact periodicity.

\subsection{Slow reasoning dynamics: the driving engine}
The reasoning dynamics $z$ evolve on the slowest timescale $\uptau_{z}$ and fully govern memory transitions through the adjacency matrix $A\in\real^{P\times P}$. While previous two-timescale Hopfield models typically invoke transitions for sufficiently large gains~\cite{HL-SP-XB:23}, the joint role of the gain parameter $\kappa$ and the initial saliency configuration $z(0)$ has not been systematically characterized. Here we provide an explicit, coefficient-wise description of the slow dynamics and derive the exact condition determining whether transitions occur or the system collapses to inactivity. In what follows, we consider a HardTanh activation function and a circulant reasoning matrix $A$. We present the main results in concise form, with extended derivations deferred to Appendix~\ref{app:slow-reas}.

Assume that the fast feature dynamics rapidly converge to a memory equilibrium, that is
\begin{equation}\label{eq:xstar}
x_{\upnu}^\star  = z_{\upnu}^2\,\upxi^{\upnu}\ \text{ with }\ z_{\upnu}^2>1.
\end{equation}
Exploiting the fast convergence of the feature layer, the slow dynamics reduce to
\begin{equation}\label{eq:zampi}
\dot z = -z + \frac{\kappa}{\sqrt{N}} A M^{\top}\Psi(x_\upnu^\star),
\end{equation}
and using (\ref{eq:xstar}) and the fact that $\Psi(x_\upnu^\star)=\upxi^{\upnu}$, \eqref{eq:zampi} can be expressed component-wise as
\begin{equation}\label{eq: z-comp2}
\begin{cases}
\dot z_{\upnu+1} = -z_{\upnu+1} + \kappa,\\
\dot z_\upmu = -z_\upmu \qquad \forall \upmu \neq \upnu+1 .
\end{cases}
\end{equation} 
Thus, all saliency weights except $z_{\upnu+1}$ decay exponentially, while $z_{\upnu+1}$ is driven by a constant input $\kappa$. More precisely, assume that at the initial time $t=0$, $z_{\upnu}(0)=Z_0>1$, while all other $z_{\upmu}(0)$, $\mu\ne \nu$ are negligible and, for simplicity, set to zero. Then, from (\ref{eq: z-comp2}), 
\begin{equation}\label{eq:z-evol}
z_\upnu(t) = Z_0 e^{-t}, \quad z_{\upnu+1}(t) = \kappa  \int_0^t e^{-(t-\uptau)} \mathrm{d}\uptau = \kappa (1 - e^{-t}).
\end{equation}
A transition from memory $\upxi^\upnu$ to $\upxi^{\upnu+1}$ occurs if two conditions are met: (i) the current equilibrium in the fast feature dynamics $x_{\upnu}^\star$ loses stability, and (ii) a stable equilibrium $x_{\upnu+1}^\star$ associated with the next memory $\upxi^{\upnu+1}$ appears. In energetic terms, condition (ii) requires the formation of a sufficiently deep well for the memory $x^{\star}_{\upnu+1}$ where the fast dynamics can roll down to, compatibly with the example in Fig.~\ref{fig:intro}.

\begin{figure}[!t]
    \centering
    \includegraphics[width=0.9\linewidth]{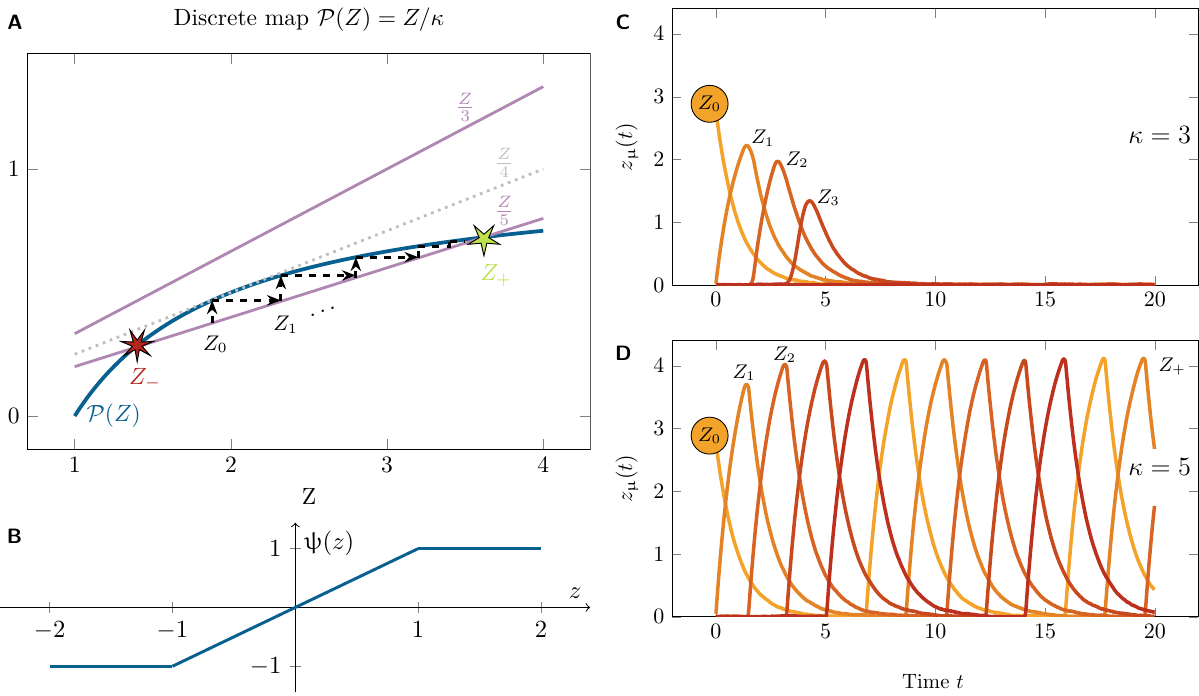}
    \caption{\textbf{Dependence of the slow reasoning dynamics on the gain $\kappa$ and the initial condition $Z_0$}. (A) Solutions of the discrete iteration $\mathcal{P}(Z)=Z/\kappa$ as a function of the gain $\kappa$. Real solutions $Z_\pm>1$ exist only for $\kappa\geq \kappa_{\text{critical}}=4$ and sequential transitions are self-sustaining only if the initial saliency weight satisfies $Z_{0}>Z_{-}$. (B) HardTanh activation function, linear in $(-1,1)$ and saturated outside this interval. (C-D) Component-wise slow reasoning dynamics for different gain regimes. (C) Subcritical gain $\kappa=3$: oscillations decay and the dynamics collapse to the origin, leading to loss of retrieval. (D) Supercritical gain $\kappa=5$: oscillations persist and entrain to the stable fixed point $Z_{+}$, enabling reliable sequential memory retrieval.}
    \label{fig: z_strength}
\end{figure}

From the stability analysis of the IDP Hopfield model, the equilibrium $x^{\star}_{\upnu}$ loses stability at the time $T_{\text{escape}}^0$ such that $Z_0e^{T_{\text{escape}}^0}=1$, that is $T_{\text{escape}}^0=\log Z_0$. In the specific case of the HardTanh, the loss of stability for the memories of the IDP Hopfield models happens as they leave the saturation regions of the activation function, hence for $z_{\upnu}^2\to 1$.
Evaluating $z_{\upnu+1}(t)$ in (\ref{eq:z-evol}) at time $T_{\text{escape}}^0$ provides the initial condition $Z_1$ for the dominant saliency weight in the next transition
$$
Z_1 = z_{\upnu+1}(T_{\text{escape}}^0)  = \kappa (1 - 1/Z_0).
$$
From the properties of the IDP Hopfield model in the case of the HardTanh activation function, the next memory equilibrium $x^{\star}_{\upnu+1}$ exists and is stable if $Z_1>1$; otherwise, the activity collapses to the origin. Iterating the above argument over successive transitions leads to the discrete-time map
\begin{equation}
Z_{t+1}=\kappa\underbrace{(1-{1}/{Z_t})}_{\mathcal{P}(Z_{t})}.
\end{equation}
which governs the peak values of the dominant saliency weights. Periodic, self-sustained transitions correspond to fixed points $Z>1$ of this map. As shown in Fig.~\ref{fig: z_strength}, such fixed points exist only for $\kappa\ge 4$, in which case they are given by
\begin{equation}
Z_{\pm} = \frac{\kappa\pm\sqrt{\kappa^2-4\kappa}}{2}>1.
\end{equation}
Moreover, for $\kappa \ge \kappa_{\text{critical}}=4$, the trajectory $\{Z_t\}_{t\ge 0}$ converges monotonically to $Z_+$ if $Z_0>Z_-$. We conclude that if $\kappa\ge 4$ and the initial condition satisfies $Z_0>Z_-$\footnote{In the case $Z_{0}<Z_{-}$ the discrete map $\mathcal{P}(Z)=Z/\kappa$ pushes $Z_{\infty}\to 1$, at which point the entire fast-slow system activity collapses to the origin.}, the system exhibits asymptotically periodic, self-sustained transitions with period $T_{\text{escape}}^\infty=\log Z_+$.

\section{Conclusions}
In this work, we developed a dynamical theory of sequential memory transitions in Hopfield-type models. By exploiting the analytical tractability of the IDP Hopfield framework, we introduced a two-layer, two-timescale retrieval–reasoning architecture that integrates fast associative recall with slow, structured transitions over learned associations. Sequentiality emerges entirely from the slow reasoning dynamics, yielding predictable memory escape times and explicit gain conditions required to sustain transitions.

In the setting of a HardTanh activation, we derived an exact gain threshold for self-sustained sequential retrieval. When this threshold is exceeded, the system exhibits stable, recurrent transitions between memories; below it, activity either collapses directly to the origin or undergoes transient, progressively weakening transitions before vanishing. These results provide a precise dynamical explanation of how sequential reasoning can arise in energy-based models without sacrificing analytical control.

Ongoing work extends this framework toward a geometric characterization of the memory manifolds supporting retrieval and of the fibers mediating transitions between them. By unifying geometric structure with timescale-separated dynamics, our aim is to establish a complete mathematical account of sequentiality in Hopfield-type networks, offering mechanistic insight into how modern energy-based architectures can support structured, multi-step reasoning.

\bibliographystyle{unsrtnat}
\bibliography{SB-main}  

\newpage

\section{Appendix}

\subsection{The IDP Hopfield model}\label{ssec: IDP_Hop}
The IDP Hopfield model presented in~\cite{BS-BG-FB-SZ:25idp} is characterized by the O.D.E.
\begin{equation}\label{eq: IDP}
    \dot x = -x + W(\upalpha)\Psi(x)
\end{equation}
where $W(\upalpha)$ is the synaptic matrix of neural interaction built from the memories of the model. The memories $\upxi^{1},\dots,\upxi^{P}$ and associated synaptic matrix $\W(\upalpha)$ obey the following assumption.
\begin{assumption}[\textbf{Memory and synaptic matrix}]\label{ass: W}
Let $P\in\mathbb{N}$ and define the set of orthogonal memories
\begin{align}\label{eq: mem}
    \Sigma &:=\{\upxi^{1},\dots,\upxi^{P}\}\subseteq\{-1,1\}^{N}\\
    {\upxi^{\upmu}}^{\top}\upxi^{\upnu}&=N\updelta_{\upmu,\upnu}\qquad \upmu,\upnu\in\{1,\dots,P\}
\end{align}
where $\updelta_{\upmu,\upnu}$ is the Kronecker delta for indices $\upmu$ and $\upnu$. For $\upalpha\in\real^{P}$ we define the input-modulated synaptic matrix $\W(\upalpha)\in\real^{N\times N}$ as
\begin{equation}\label{eq: W(a)}
    \W(\upalpha)=\frac{1}{N}\sum_{\upmu=1}^{P}\upalpha_{\upmu}\upxi^{\upmu}{\upxi^{\upmu}}^{\top}.
\end{equation}
\end{assumption}
The orthogonal memories constraint is necessary to derive rigorous results on existence and stability of fixed points, but can be relaxed to orthogonality in expectation in implementations.
\begin{equation}
    \mathbb{P}[\upxi^{\upnu}_{i}=1]=\mathbb{P}[\upxi^{\upnu}_{i}=-1]=\frac{1}{2}\qquad \forall\upnu=1,\dots,P,\quad \forall i=1,\dots,N
\end{equation}
\begin{remark}
    Notice that by constructing the memory matrix
    \begin{equation}
        M=\frac{1}{\sqrt{N}}[\upxi^{1},\dots,\upxi^{P}]
    \end{equation}
    we can rewrite the input-modulated synaptic matrix as
    \begin{equation}
        W(\upalpha)=M\diag(\upalpha)M^{\top}
    \end{equation}
    The canonical synaptic matrix $W$ of the Hopfield model is a specific instance of~\eqref{eq: W(a)} for the choice $\upalpha=\vectorones$ (or a scalar multiple of it).
\end{remark}
We now specialize the treatment of the IDP Hopfield model as presented in~\cite{BS-BG-FB-SZ:25idp} to the case of the HardTanh activation function. Specifically, we consider $\Psi:\real^N\to\real^N$ to be diagonal and homogeneous, i.e.,
\begin{equation}
    \Psi_i(x)=\uppsi(x_i)
\end{equation}
with $\uppsi(s)=\max\{-1,\min\{1,s\}\}$ for $s\in\real$.

\subsubsection{Existence of the memories as equilibrium points}

Starting from the condition for the existence of the equilibria in~\cite{BS-BG-FB-SZ:25idp}, we gain an explicit parametrization of $x^{\star}_{\upnu}=\upgamma_{\upnu}\upxi^{\upnu}$ when such equilibria exist. In particular, the saliency weights $\upalpha_{\upnu}=z^2_\upnu$, for $\upnu=1,\dots,P$, must uniquely satisfy
\begin{equation}\label{eq: constr-z}
    \upalpha_{\upnu}=z^{2}_{\upnu}=\frac{\upgamma_{\upnu}}{\uppsi(\upgamma_{\upnu})}
\end{equation}
which in this specific case becomes
\begin{equation}z_{\upnu}^2=
    \begin{cases}
        1\qquad\ \ \upgamma_{\upnu}\in(0,1)\\
        \upgamma_{\upnu}\qquad \upgamma_{\upnu}\in[1,+\infty)
    \end{cases}
\end{equation}
If $z_{\upnu}^2<1$, then there does not exist $\upgamma_{\upnu}>0$ such that~(\ref{eq: constr-z}) is satisfied. Instead, for $z_{\upnu}^{2}=1$~(\ref{eq: constr-z}) is trivially satisfied by any $\upgamma_{\upnu}\in(0,1]$. Finally, when $z_{\upnu}^2>1$ it is easy to see that no $\upgamma_{\upnu}\in(0,1]$ satisfies~(\ref{eq: constr-z}), and that for any $\upgamma_{\upnu}>1$ we have
\begin{equation}
    z_{\upnu}^2=\upgamma_{\upnu}
\end{equation}
It then follows that existing equilibria are explicitly parametrized as $x^{\star}_{\upnu}=z^{2}_{\upnu}\upxi^{\upnu}$. 

\subsubsection{Local stability of the memories as equilibrium points}

The local stability condition for the memories requires a direct computation of the Jacobian of the Hopfield map $f(x,\upalpha)=-x+W(\upalpha)\Psi(x)$\footnote{Notice that by the non-smoothness of the HardTanh activation function, the Jacobian is not defined for any $x\in\partial[-1,1]^N$ boundary of the hypercube $[-1,1]^N$. Since the existence of the memories requires $z_{\upnu}^2>1$, then we automatically have that $x^\star_\upnu\notin \partial[-1,1]$, and for the present treatment with do not consider singularities on measure zero sets.}. At the equilibrium points, we have
\begin{align}
    \frac{\partial f}{\partial x}(x^{\star}_{\upnu}) &= -\displaystyle \mI + W(\upalpha)\frac{\partial\Psi}{\partial x}(x^{\star}_{\upnu})\nonumber\\
    &=-\mI + W(\upalpha)\uppsi'(z_{\upnu}^{2})\mI
\end{align}
where in the last passage we have exploited the symmetry of the derivative of the HardTanh with respect to the origin. In particular, we will have that $\uppsi'(z_{\upnu}^2)=0$ for any $z_{\upnu}^2>1$. Thus
\begin{equation}\label{eq: stab}
    \frac{\partial f}{\partial x}(x^{\star}_{\upnu})=-\mI\qquad \forall z_{\upnu}^2>1
\end{equation}
and therefore the existence condition for the memories and their local stability condition coincide. In particular, we conclude that $x_{\upnu}^{\star}$ exists and is a locally exponentially stable memory of the IDP Hopfield model with HardTanh activation function for all $z_{\upnu}^2>1$.

\subsection{Slow reasoning dynamics: under the hood}\label{app:slow-reas}

Under the fast relaxation of the feature dynamics to the stable equilibrium $x^{\star}_{\upnu}=z_{\upnu}^2\upxi^\upnu$, we want to study the unfolding of the slow dynamics. Starting from the definition of the slow dynamics, and noticing that for HardTanh activation function $\Psi(x^{\star}_{\upnu})=\uppsi(z^{2}_{2})\upxi^{\upnu}=\upxi^\upnu$, since $z_{\upnu}^2>1$, we get
\begin{align}
    \dot z &= -z+\frac{\kappa}{\sqrt{N}}AM^{\top}\Psi(x^{\star}_{\upnu})\nonumber\\
           &= -z+\frac{\kappa}{\sqrt{N}}AM^{\top}\upxi^{\upnu}\nonumber\\
           &= -z+\kappa A\displaystyle \ve^{(\upnu)}\nonumber\\
           &= -z+\kappa \displaystyle \ve^{(\upnu+1)},
\end{align}
where $\ve^{(\ell)}$ denotes the $\ell$-th vector of the canonical basis of $\mathbb{R}^P$.
It then becomes clear that the study of the dynamics is reduced to the simple components condition
\begin{equation}
    \begin{cases}
        \dot z_{\upnu+1}=-z_{\upnu+1}+\kappa,\\
        \dot z_{\upmu}=-z_{\upmu},\qquad \upmu\neq\upnu+1
    \end{cases}
\end{equation}
and considering for simplicity initial conditions $z_{\upnu}(0)=Z_{0}>1$, $z_{\upmu}(0)=0$ for all $\upmu\neq \upnu$ and integrating we get
\begin{equation}\label{eq: scal-sal}
    \begin{cases}
        z_{\upnu}(t)=Z_{0}e^{-t}\\
        z_{\upnu+1}(t)=\kappa\int_{0}^{t}e^{-(t-\uptau)}\ d\uptau=\kappa(1-e^{-t})
    \end{cases}
\end{equation}
In order for the transition to occur, we need the memory $x^{\star}_{\upnu}$ to lose stability as given by~(\ref{eq: stab}). Therefore, there must exist a first escape time $T^{0}_{\text{escape}}>0$ such that
\begin{equation}
    1=Z_{0}e^{-T^{0}_{\text{escape}}}
\end{equation}
which trivially gives $T^{0}_{\text{escape}} = \log Z_{0}$. Using the new estimate of the first escape time in~(\ref{eq: scal-sal}) we obtain that
\begin{align}
    z_{\upnu+1}(T^{0}_{\text{escape}})&=\kappa(1-e^{-\log Z_{0}})\nonumber\\
    &=\kappa\left(1-\frac{1}{Z_{0}}\right).
\end{align}
Calling $Z_{1}=z_{\upnu+1}(T_{\text{escape}}^0)$, we will have a new escape time for the memory $x^{\star}_{\upnu+1}$ estimated as $T^1_{\text{escape}}=\log Z_1$, from which it will be possible to estimate $Z_{2}=\kappa(1-1/Z_{1})$ and so forth. This iterative mechanism yields the one-dimensional discrete-time dynamical system
\begin{equation}\label{eq:tmap}
Z_{t+1}=\kappa\left(1-\frac{1}{Z_t}\right)=\kappa \mathcal{P}(Z_{t}):=\mathcal F(Z_t).
\end{equation}
Fixed points $Z>1$ of (\ref{eq:tmap}) correspond to periodic saliency peaks of the components of $z$, i.e., to periodic self-sustained transitions. They are the solutions of $\mathcal F(Z)=Z$, equivalently the real roots of
\begin{equation}
Z^2-\kappa Z+\kappa=0.
\end{equation}
Such roots are real and greater than $1$ if and only if \(\kappa\ge 4\), in which case
\begin{equation}
Z_{\pm}=\frac{\kappa\pm\sqrt{\kappa^2-4\kappa}}{2}.
\end{equation}

To determine convergence, we examine the increment
$$
Z_{t+1}-Z_t=\mathcal F(Z_t)-Z_t,
$$
whose sign indicates whether the sequence increases or decreases.
A direct computation gives
$$
\mathcal F(Z)-Z
=\kappa\left(1-\frac{1}{Z}\right)-Z
=-\frac{Z^2-\kappa Z+\kappa}{Z}
=-\frac{(Z-Z_-)(Z-Z_+)}{Z}.
$$
For $Z>1$, the denominator is positive, so the sign of
$\mathcal F(Z)-Z$ is opposite to that of $(Z-Z_-)(Z-Z_+)$.

\medskip
\noindent
\emph{Case $\kappa>4$.}
The fixed points are distinct and satisfy $1<Z_-<Z_+$. Then
$$
\mathcal F(Z)-Z
\begin{cases}
<0, & 1<Z<Z_-,\\
>0, & Z_-<Z<Z_+,\\
<0, & Z>Z_+.
\end{cases}
$$
Hence, if $1<Z_0<Z_-$ the sequence decreases toward the boundary $Z=1$ 
and transitions are not sustained. 
If $Z_0>Z_-$, the trajectory is monotone and remains trapped between 
$Z_-$ and $Z_+$, converging to $Z_+$. 

\medskip
\noindent
\emph{Case $\kappa=4$.}
The two fixed points coincide at $Z_\pm=2$, and
\[
\mathcal F(Z)-Z
=-\frac{(Z-2)^2}{Z}\le 0.
\]
Hence $Z_{t+1}\le Z_t$ for all $Z>1$, with equality only at $Z=2$. 
If $Z_0>2$, the sequence decreases monotonically to $2$. 
If $1<Z_0<2$, it decreases away from $2$ toward the boundary $Z=1$, 
so sustained transitions do not occur. 

In conclusion, for $\kappa\ge 4$ and $Z_0>Z_-$, the model supports asymptotically periodic, self-sustained transitions, and the corresponding asymptotic period is given by
\begin{equation}
T^{\infty}_{\mathrm{escape}}=\log Z_+.
\end{equation}

\end{document}

%% file: math_commands.tex
\usepackage{amsmath,amsfonts,bm}

\def\eqref#1{equation~\ref{#1}}

\def\1{\bm{1}}

\def\ve{{\bm{e}}}

\def\mI{{\bm{I}}}

\DeclareMathAlphabet{\mathsfit}{\encodingdefault}{\sfdefault}{m}{sl}
\SetMathAlphabet{\mathsfit}{bold}{\encodingdefault}{\sfdefault}{bx}{n}

